\documentclass[runningheads]{llncs}

\usepackage{svg}
\usepackage{array}
\usepackage{amsmath}
\usepackage{amsfonts}
\usepackage{graphicx}
\usepackage{tabularx}
\usepackage{booktabs}
\usepackage{multirow}
\usepackage{colortbl}
\usepackage{enumitem}
\usepackage{hyperref}
\usepackage[misc]{ifsym}
\usepackage[ruled,linesnumbered]{algorithm2e}

\begin{document}

\title{MTFusion: Reconstructing Any 3D Object from Single Image Using Multi-Word Textual Inversion}

\titlerunning{MTFusion}

\author{Yu Liu \inst{1} \and
Ruowei Wang \inst{2} \and
Jiaqi Li \inst{1} \and
Zixiang Xu \inst{2} \and
Qijun Zhao \inst{1,2}\textsuperscript{(\Letter)}
}

\authorrunning{Y. Liu et al.}

\institute{National Key Laboratory of Fundamental Science on Synthetic Vision, Sichuan University, Chengdu, China \and
College of Computer Science, Sichuan University, Chengdu, China
\email{qjzhao@scu.edu.cn}}

\maketitle

\begin{abstract}
Reconstructing 3D models from single-view images is a long-standing problem in computer vision.
The latest advances for single-image 3D reconstruction extract a textual description from the input image and further utilize it to synthesize 3D models.
However, existing methods focus on capturing a single key attribute of the image (\textit{e.g.}, object type, artistic style) and fail to consider the multi-perspective information required for accurate 3D reconstruction, such as object shape and material properties.
Besides, the reliance on Neural Radiance Fields hinders their ability to reconstruct intricate surfaces and texture details.
In this work, we propose MTFusion, which leverages both image data and textual descriptions for high-fidelity 3D reconstruction.
Our approach consists of two stages.
First, we adopt a novel multi-word textual inversion technique to extract a detailed text description capturing the image's characteristics.
Then, we use this description and the image to generate a 3D model with FlexiCubes.
Additionally, MTFusion enhances FlexiCubes by employing a special decoder network for Signed Distance Functions, leading to faster training and finer surface representation.
Extensive evaluations demonstrate that our MTFusion surpasses existing image-to-3D methods on a wide range of synthetic and real-world images.
Furthermore, the ablation study proves the effectiveness of our network designs.

\keywords{3D Reconstruction \and Diffusion Model \and Textual Inversion}
\end{abstract}
\section{Introduction}
Reconstructing 3D objects from single-view images \cite{3drec_gc,3drec_tac,3drec_de} remains a fundamental challenge in computer vision.
Traditionally, tackling this problem with neural networks requires vast amounts of 3D training data \cite{ws3drecon,bird3drecon,3dfr}, which is a significant bottleneck hindering the reconstruction progress.
However, the recent advances in diffusion models \cite{ddpm,ddim,ldm}, particularly their success in conditional generation tasks such as text-to-image generation \cite{sjc,dreamfusion,magic3d,prolificdreamer,fantasia3d,dreamgaussian}, presents a novel approach for solving this ill-posed problem.
Given a pretrained text-to-image diffusion model, DreamFusion \cite{dreamfusion} proposes Score Distillation Sampling (SDS) to utilize it as the knowledge prior for guiding the training of a Neural Radiance Field (NeRF) \cite{nerf}, making it possible to generate arbitrary 3D object from textual descriptions, even without any explicit 3D training data.

\begin{figure}
    \centering
    \includegraphics[width=\textwidth]{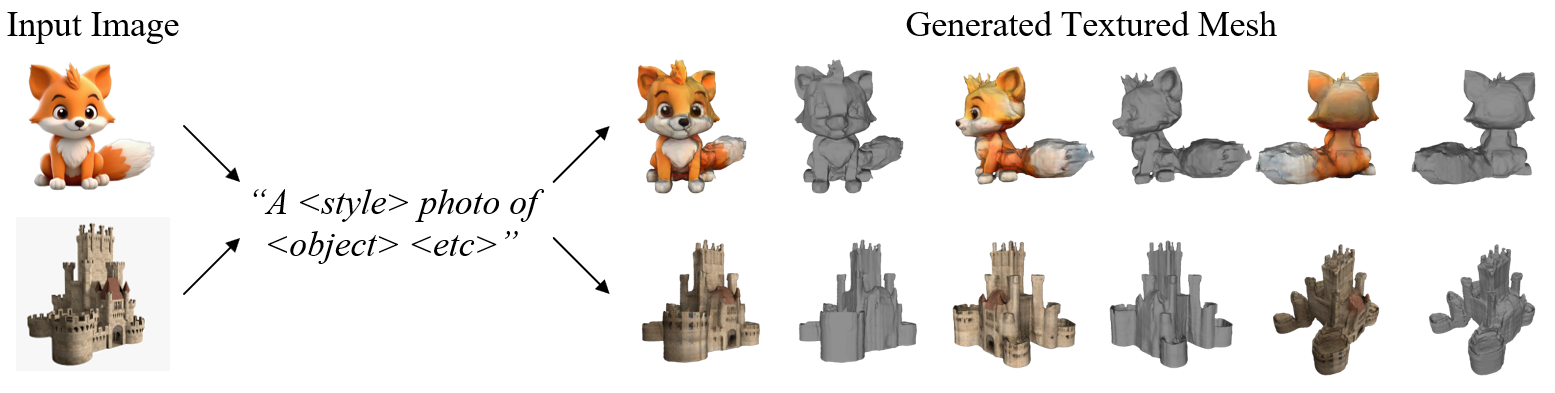}
    \caption{Given a single-view image, MTFusion generates a textured mesh using supervision from the image and a pseudo-prompt.}
    \label{fig:teaser}
\end{figure}

This approach leads to a naturally occurring question: can text-to-3D generation be extended to image-to-3D generation by mapping the input image to the text embedding space?
The key point of this strategy lies in finding a text embedding that could faithfully reconstruct the given image.
Some attempts have been made by recent works \cite{makeit3d,realfusion,magic123}.
Specifically, Make-it-3D \cite{makeit3d} utilizes a pretrained image caption model to get a textual description for the input.
RealFusion \cite{realfusion} introduces textual inversion for finding the optimal word embedding corresponding to the input in the latent space of a pretrained text encoder.
Afterward, we can generate a 3D object by conditioning the NeRF on the input image and the text.
This effectively utilizes the pretrained diffusion model's prior knowledge to guide the reconstruction process.
In essence, the above methods both consider using textual descriptions to serve as a bridge, which provides missing information in the input view, to mitigate the gap between 2D image and reconstructed 3D object.

Vanilla textual inversion \cite{ti} hinges on the utilization of single-word embedding to encapsulate the visual characteristics gleaned from the input image.
This approach has garnered significant recognition for its remarkable ability to capture these attributes effectively.
However, when seamlessly integrated into an image-to-3D reconstruction framework, the limitation of single-word embedding emerges as it tries to describe only one key attribute of the image (\textit{e.g.}, object type, artistic style), neglecting the need for multi-perspective information needed for 3D reconstruction.

Another area of concern is that current prevailing 3D generation methods \cite{sjc,dreamfusion,magic3d,realfusion} have conventionally leaned on the implicit scene modeling paradigm offered by NeRFs \cite{nerf,mip_nerf,instantngp}.
However, when it comes to reconstructing intricate surfaces and capturing their texture details, NeRF-based methods fall short due to their coupling of geometry and appearance.
Furthermore, the limited portability and editability of NeRFs also prevent their wider applications in 3D modeling.
In response to these limitations, researchers have actively explored alternative 3D representations.
Hybrid 3D representations \cite{dmtet,flexicubes} have emerged as a promising avenue for surpassing the capabilities of NeRFs.
These approaches advocate for explicitly modeling the surface of the object, enabling a more targeted reconstruction process.
With the utilization of DMTet \cite{dmtet}, several methods \cite{fantasia3d,magic123,get3d} achieve superior reconstruction of both the geometric details and material properties.
Nevertheless, the vertices of DMTet are unable to move independently, limiting its representational ability in surface modeling.

Driven by the above analysis, we introduce MTFusion for reconstructing 3D objects from single-view images. 
Our method leverages both visual and textual supervision to achieve high-fidelity reconstruction.
It consists of two stages: textual inversion and text-to-3D modeling.
The first stage focuses on extracting a latent textual description that captures the characteristics of the input image.
To achieve this, we propose multi-word textual inversion, which is a novel technique for finding the optimal textual description of the image.
Specifically, we initialize a prompt template, which incorporates pseudo text embeddings to capture visual attributes like object type and style within the input image, and then we optimize this prompt to make them more semantically consistent.
Different from vanilla textual inversion, we employ a gradient-free optimization strategy to refine the embedding, which significantly reduces computational costs by circumventing the need for gradient information in the diffusion model.

The second stage utilizes the extracted textual embedding, along with the input image, to generate a 3D model.
We adopt FlexiCubes as the underlying 3D representation for modeling.
However, directly integrating FlexiCubes into our framework leads to slow training speed and generative artifacts due to the score distillation sampling scheme.
To overcome this issue, we propose an SDF decoder network in conjunction with hashgrid positional encoding, for efficiently extracting the FlexiCubes parameters from grid points. 
This approach empowers the network to converge faster on an accurate representation of the 3D object and significantly enhances the representational ability of the neural fields employed by MTFusion.
The generated mesh is further incorporated into a texture network for appearance reconstruction.
Evaluations on both synthetic and real images confirm MTFusion's dominance in single-image 3D reconstruction compared to baseline methods.
Furthermore, ablation studies validate the critical contributions of both the multi-word textual inversion and the enhanced FlexiCubes representation within our framework.

The contributions of this paper can be summarized as:
\begin{itemize}
    \item We propose MTFusion for reconstructing arbitrary 3D object from single image using multi-word textual inversion.
    \item By incorporating hashgrid positional encoding into FlexiCubes, we improve the generation stability and training speed of SDS-based 3D modeling.
    \item Our experiments demonstrate MTFusion's ability for high-fidelity 3D reconstruction. A broad range of objects generated exhibit intricate geometric details and realistic textures that closely adhere to the reference image.
\end{itemize}

\section{Related Work}
\subsection{Textual Inversion}
The inversion of diffusion model \cite{dmi} seeks to unearth the underlying noise map and conditioning vector corresponding to a given image. 
\cite{ti} proposes textual inversion as a method for pretrained text-to-image diffusion models to represent concepts, such as objects or artistic styles, which might be challenging to describe using natural language.
However, previous approaches \cite{ti,mtti,dmi} necessitate backpropagation through a pretrained diffusion model to adjust embedding parameters, imposing high computational demands.
The advent of gradient-free techniques \cite{gfti} boosted computational efficiency by utilizing evolutionary strategies for continuous optimization, circumventing the need for gradient calculation.
In 3D generation, leveraging textual inversion for semantic token generation from a single image has set new benchmarks\cite{realfusion,magic123}.
Advances in distributional prompt learning \cite{dreamdist} further refine this by generating personalized images through the learned prompt distribution, offering enhanced semantic consistency and stylistic variation, thereby broadening the applicability and flexibility of text-to-image models.

\subsection{Differentiable Rendering}
Numerous prior endeavors \cite{mc,dc,dualmc,dmc,dmtet,flexicubes,3dsst} in 3D modeling have culminated in the acquisition of a definitive explicit mesh.
This is typically accomplished by exporting the Neural Radiance Fields (NeRF) \cite{nerf} representation into a mesh-based counterpart through surface extraction techniques.
Among these methods, Marching Cubes (MC) \cite{mc} has emerged as the most widely adopted approach.
However, MC suffers from inherent topological ambiguities and cannot guarantee the faithful representation of sharp features in the resulting mesh.
An alternative technique, known as Dual Contouring (DC) \cite{dc}, leverages a dual representation where mesh vertices are extracted per cell, and the estimation of vertex position relies on local details, aiming to address some of the limitations posed by MC.

In recent years, the intersection of machine learning and 3D mesh reconstruction has yielded exciting developments.
Researchers have delved into gradient-based mesh reconstruction schemes \cite{nvdiffrast,nvdiffrec}, which operate by extracting surfaces from implicit functions encoded via convolutional neural networks.
These schemes evaluate objectives directly on the mesh, allowing for utilization in learning-based methods.
One notable advancement is DMTet \cite{dmtet}, which introduces a differentiable marching tetrahedra layer that bridges the gap between implicit and explicit representations.
By converting the implicit signed distance function to an explicit mesh, DMTet retains the ability to effectively represent complex topology.
However, the vertices of the extracted mesh in DMTet lack independent mobility, resulting in surface artifacts.
To address this challenge, FlexiCubes \cite{flexicubes} emerges as a promising solution.
Building upon the foundation of Dual Marching Cube extraction, FlexiCubes introduces additional weight parameters to enhance the flexibility of the mesh-based representation.
These gradient-based surface extraction methods not only accommodate arbitrary topology due to their combination with implicit functions but also offer an end-to-end training framework, facilitating seamless integration into broader applications.
\section{Method}
\begin{figure}
    \centering
    \includegraphics[width=\textwidth]{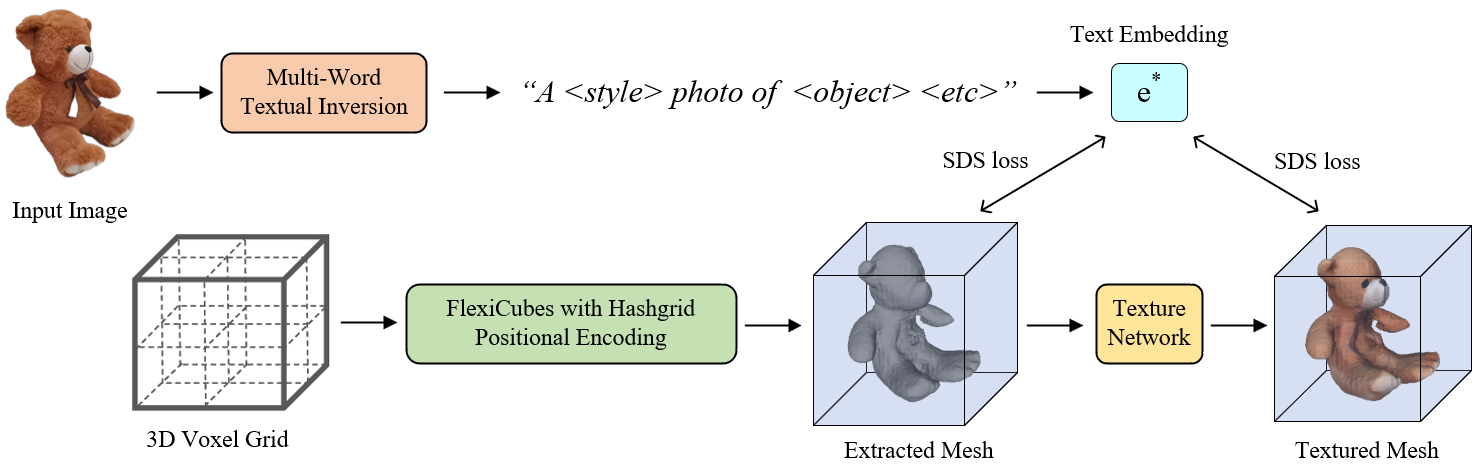}
    \caption{Overview of MTFusion. Our approach extracts a textual description from the input image and constrains the 3D model based on this description and the image.}
    \label{fig:pipeline}
\end{figure}

In this section, we introduce our method of MTFusion for 3D object reconstruction from a single image.
As shown in Fig. \ref{fig:pipeline}, instead of relying on multiple views of the object, we harness the underlying text descriptions associated with the image to compensate for the missing details.
To achieve this, we propose to use multi-word textual inversion to extract the information of diverse aspects from the input image, then condition the 3D model on both 2D and 3D objectives to get the final textured mesh.
Specifically, we first introduce some preliminaries in sec. \ref{sec:preliminary}, and then the two main training stages of our method: textual inversion in sec. \ref{sec:fitst_stage} and 3D reconstruction in sec. \ref{sec:second_stage}

\subsection{Preliminary}
\label{sec:preliminary}
\subsubsection{Latent Diffusion Models.} LDMs \cite{ldm} are machine learning models designed to learn the underlying structure of a dataset by mapping it to a lower-dimensional latent space. 
It involves two fundamental processes: a forward diffusion process that progressively obscures the original image and a reverse denoising process that recovers the original image from its noisy counterpart. 
The forward process gradually adds noise to an input image over a series of discrete time steps.
To be specific, consider an image denoted as $I$.
At each time step $t$, a fraction of noise is added, controlled by the parameter $\sigma_t$. 
The noisy image at time $t$ can be expressed as $I_t=\alpha_tI+\sigma_t\epsilon$, where $\sigma_t$ represents the standard deviation of the noise added at time $t$ and $\epsilon$ represents the sample drawn from a Gaussian distribution with zero mean and unit variance.
Additionally, the variance of the noise, $\sigma_t^2$, is related to $\alpha_t$ as $\sigma_t^2 = 1 - \alpha_t^2$.

During the reverse process, a denoising neural network is used.
Given the noisy image $I_t$ and the noise level $t$, the network predicts the noise component $\epsilon$ as $\hat{\epsilon}=\Phi(I_t;t)$. 
By iteratively applying this calculation, we can reconstruct the original image $I$ from the fully noised version $I_0$.
The diffusion model is usually trained on a large collection of images by minimizing the LDM loss:
\begin{equation}
    \mathcal{L}_{\text{LDM}} (\Phi ; \mathcal{D}) = \frac{1} {|\mathcal{D}|} \sum_{I \in \mathcal{D}} ||\Phi(\alpha_tI+\sigma_t \epsilon ; t, y) - \epsilon||^2,
    \label{equ:ldm}
\end{equation}
where $\mathcal{D} = \{I\}$ indicates the training dataset. 
LDMs can be readily expanded to generate samples from a distribution that is conditioned on a given prompt. 
This conditioning process is achieved by introducing the prompt as an additional input to the diffusion model's noise predictor, enabling LDMs for downstream tasks like image processing and generative modeling.

\subsubsection{Score Distillation Sampling.} SDS \cite{sjc,dreamfusion} emerges as a novel technique that bridges the gap between text prompts and tangible 3D models. 
The crux of this approach lies in its ability to extract a comprehensive 3D rendition from a 2D diffusion model, denoted as $p(I|e)$, and a given prompt $e$. 
The optimization process involves refining the randomly initialized 3D model through gradient descent using a pretrained 2D diffusion model as guidance, and the objective is to minimize the loss between the 2D renderings of the 3D model from various angles and the given prompt $e$, detailed as:
\begin{equation}
    \bigtriangledown_{\sigma,c}\mathcal{L}_{SDS}(\sigma,c;\pi,e,t) = E_{t,\epsilon}[w(t)(\Phi(\alpha_tI + \sigma_t\epsilon;t,e) - \epsilon)\cdot\frac{\partial I}{\partial (\sigma,c)}],
    \label{equ:sds}
\end{equation}
where $I = R(\sigma, c, \pi)$ is the image rendered from a randomly-sampled viewpoint $\pi$ and $w(t)$ is a weighting function depending on timestep $t$. 
Remarkably, score distillation sampling circumvents the need for labeled 3D training data.
It leverages the pretrained 2D diffusion models as effective priors for 3D synthesis.

\begin{figure}
    \centering
    \includegraphics[width=0.95\textwidth]{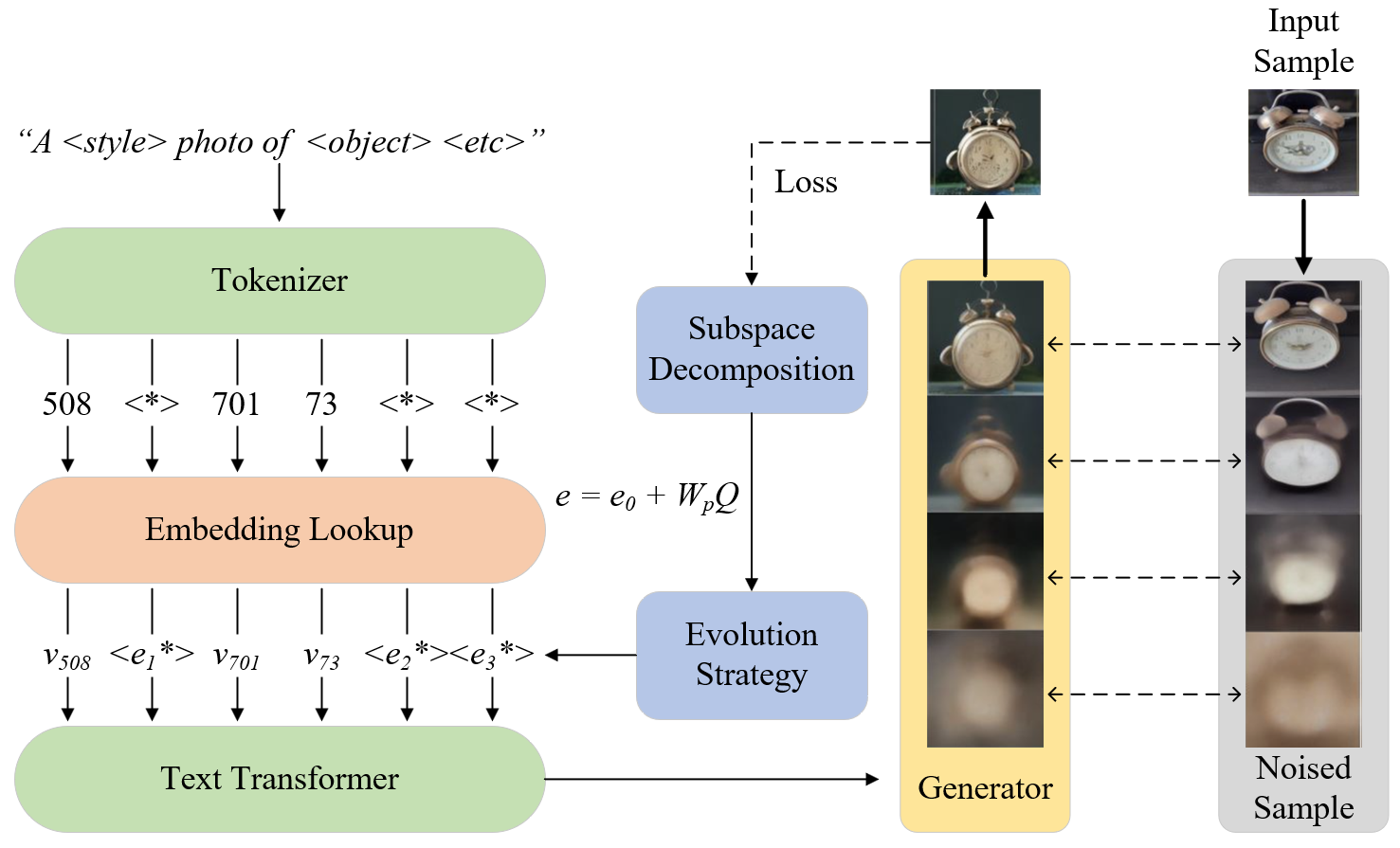}
    \caption{Overview of our proposed Multi-Word Textual Inversion. The optimization of text embedding is based on a gradient-free approach, which iteratively employs an evolution strategy to explore and exploit pseudo-token embeddings.}
    \label{fig:mwti}
\end{figure}

\subsection{First Stage: Multi-Word Textual Inversion}
\label{sec:fitst_stage}
This stage's main purpose is to obtain a pseudo-textual description of the input image for further text-to-3D synthesis using score distillation sampling.
As shown in Fig. \ref{fig:mwti}, given an image and a pretrained diffusion model, we aim to extract the feature from pixel space to text-embedding space by exploring the latent space of the text encoder to find the optimal embedding $e^*$ that could reconstruct the input.
Our approach centers around the strategic engineering of a specialized text prompt, denoted as $e_0$, specifically tailored for input image $I_0$. 
This prompt serves as a surrogate for multi-view information, approximating the conditional distribution $p(I|I_0)$.
Our objective is to minimize the diffusion loss, as defined in Equation \ref{equ:ldm}, with respect to the prompt $e_0$. 
Importantly, during this optimization process, all other parts of text embeddings within the prompt and model parameters remain fixed.

Different from the vanilla textual inversion, which sets the prompt template as \textit{"an image of $<$e$>$"}, we draw inspiration from \cite{mtti}. 
Specifically, our prompt formulation is derived from \textit{"a $<$style$>$ image of $<$object$>$ $<$etc$>$"}, where $<$style$>$, $<$object$>$ and $<$etc$>$ represents the novel token for describing the overall visual style, salient object and residual information of the image respectively, \textit{e.g.} \textit{"a $<$photo-realistic$>$ image of $<$a horse$>$ $<$running$>$"}, \textit{"a $<$cartoon$>$ image of $<$a toy$>$ $<$on the beach$>$"}.
Each token focuses on one aspect of the information contained in the image $I_0$, aiming to describe a feature that is impractical to translate into natural language. 
Afterward, these tokens are introduced into the vocabulary of the text encoder within our diffusion model, eliminating the need for an explicit caption \cite{makeit3d} or description of $I_0$.

\subsubsection{Optimization Strategy.}
Our optimization strategy is based on the recently proposed gradient-free approach \cite{gfti}. 
Due to concerns about computational efficiency, we aim to determine the optimal pseudo-token embedding $e*$ in a gradient-free scenario, which significantly reduces the duration of the textual inversion stage.
To tackle this high-dimensional optimization problem posed by the text embedding $e$, we utilize a decomposition strategy to denote $e$ as the sum of a well-initialized embedding $e_0$ and an incremental component $W_pQ$, with $W_p$ and $Q$ indicating the projection matrix and corresponding coefficient respectively.
Thus, our objective function becomes:
\begin{equation}
    Q^*=arg\min_{Q\in Q^{\prime}} L(e_0+W_pQ)
    \label{equ:q}
\end{equation}
where $Q^{\prime}$ represents the sub search space. Therefore, the training process comprises two steps:
\begin{itemize}
    \item Initialization: We initialize the pseudo-token embedding $e_0$ using the CLIP-similarity comparison strategy.
    \item Optimization: We iteratively evolve the incremental component $Q$ using a linear projection matrix $W_p$. 
    To ensure accurate optimization, we fix the added noise level (parameter $t$) during each evaluation.
\end{itemize}

\paragraph{Initialization.}
For the initialization of tokens in the prompt template, we replace the token with word embeddings from the vocabulary of the pretrained text encoder and then compute the CLIP similarity score between every pair of visual and text embeddings. 
Subsequently, the scores are used as embedding weights to get the initialization embedding of the pseudo-token. 
The pseudo-token for object and style are initialized separately, and the pseudo-token for residual information is initialized randomly.

\paragraph{Subspace Decomposition.}
To reduce the computational complexity, we utilize Principal Components Analysis (PCA) for subspace decomposition of the high-dimensional word embeddings.
Our approach involves collecting all embeddings from the vocabulary of the text encoder as training data. 
Subsequently, we compute the PCA on this dataset, yielding a set of $d$ eigenvectors. 
These eigenvectors serve as the foundation for constructing the projection matrix $W_p$. 
This methodology allows us to inverse-transform the optimized incremental components back into the original space, guided by well-defined rules and linear transformations. 

\paragraph{Evolution Strategy.}
To avoid the time-consuming process of gradient calculation, we adopt the evolution algorithm \cite{es} to search the text encoder's latent space. 
The Covariance Matrix Adaptation Evolution Strategy (CMA-ES) \cite{cma_es} stands as a prominent evolutionary algorithm employed for non-convex gradient-free optimization in continuous domains.
Specifically, CMA-ES maintains a multivariate normal distribution to guide the search towards promising solution regions within the latent space.
Its versatility and effectiveness have made it a go-to choice for tackling complex optimization problems.

\subsection{Second Stage: SDS-Based 3D Modeling}
\label{sec:second_stage}
The goal of this stage is to generate a 3D model based on the input image $I_0$ and the obtained text embedding $e*$.
We achieve this by conditioning the 3D model on image reconstruction loss with $I_0$ and SDS loss with $e^*$.
In other words, it is a text-to-3D generation process additionally constrained by 2D images.
We propose to use the hybrid 3D representation of FlexiCubes as our modeling tool.
Due to the disentanglement of geometry and appearance property of FlexiCubes, we divide this stage into two progressive steps: geomotry reconstruction and texture reconstruction, detailed as follows.

\subsubsection{Geometry Reconstruction.}
Text-to-3D approaches in the past \cite{sjc,dreamfusion,magic3d} usually employ neural radiance fields \cite{nerf,mip_nerf,instantngp} as the implicit scene representation, intertwining geometry with color and appearance, as well as relying on post-processing methods like Marching Cubes for surface extraction.
Due to NeRFs' limitations in surface reconstruction, some existing works utilize DMTet, which is a hybrid representation that benefits from both implicit function and explicit 3D mesh.
However, the vertices of DMTet cannot be freely placed in the grid, leading to suboptimal results when attempting to create high-fidelity 3D surfaces by back-propagating language supervision through pretrained text-to-image models.
Inspired by \cite{fantasia3d}, we propose to use FlexiCubes \cite{flexicubes} as the underlying 3D representation of our framework.

\paragraph{Surface Extraction with FlexiCubes.}
Based on differentiable rendering, FlexiCubes \cite{flexicubes} is a kind of hybrid 3D representation that combines the benefits of both implicit and explicit representations. 
Besides SDF values and vertex position offsets, three types of per-cube weight parameters are used in optimizing the underlying scaler field: interpolation weights for positioning dual vertices in space, splitting weights for controlling how quadrilaterals are split into triangles, and deformation vectors for spatial alignment. 
Based on these parameters, a triangle mesh can be obtained via Dual Marching Cubes \cite{dualmc}.

\paragraph{SDF Network.}
However, we found that directly using FlexiCubes as the neural network parameters for score distillation sampling brings the problem of slow training speed and generative artifacts. 
Inspired by \cite{instantngp,fantasia3d}, we propose to use an MLP with multi-resolution hash grids positional encoding to predict SDF values, vertex position offsets, and additional weights for FlexiCubes. 
This combination significantly boosts the representation power of neural fields and leads to a remarkably faster convergence speed. 
Additionally, we adopt a coarse-to-fine training strategy for geometric reconstruction. 
During the coarse training stage, we first initialize the SDF network for FlexiCubes as an ellipsoid. 
Then, we utilize the downsampled latent code of normal map and mask in diffusion training under the supervision of input image and text embedding, facilitating rapid updates to the shape representation. 
During the refining stage, only the high-resolution normal map was used to generate the finer surface details.
The overall training objective during geometry reconstruction is (1) 2D reconstruction loss at reference view and (2) SDS loss (Equation \ref{equ:sds}) at novel views.

\subsubsection{Appearance Reconstruction.}
In the context of texture synthesis, following \cite{nvdiffrec,nvdiffrast}, we adopt an approach involving a texture neural field that is seamlessly integrated with an extracted mesh, allowing us to obtain a high-quality textured surface.
Specifically, given a camera viewpoint, we perform rasterization on the mesh, transforming it into a 2D image. which is then meticulously compared against the ground-truth reference. 
We leverage a texture neural network to compute per-pixel linearly-interpolated surface positions, surface normals, and view direction. 
The surface positions are further encoded using a multi-resolution hashgrid, serving as the input to the texture network along with predicted normals and view directions.
The texture network, which we implemented as a shallow multi-layer perceptron, predicts per-pixel RGB values for the rendered image. 
The modeling of appearance is also constrained by reconstruction loss at the reference view and SDS loss (Equation \ref{equ:sds}) at novel views.
\section{Experiments}
\subsection{Implementation Details}
For our diffusion model prior, we employ the open-source Stable Diffusion v1.5 \cite{ldm}, which is pretrained on the LAION dataset \cite{laion}, as the backbone. 
The SDF decoder network uses a simple MLP with a skip connection and 64 hidden units.
The texture network is implemented as a three-layer MLP with 256 hidden units. 
During geometry modeling, We set the grid resolution as 128 for the consideration of balance between speed and quality.
All of our experiments are conducted on a single 24GB RTX 4090 GPU, the entire reconstruction process for an image takes approximately one hour.

\subsection{Qualitative Experiment}
To gauge the performance of MTFusion within the context of image-to-3D reconstruction, we compare our method against recent state-of-the-art SDS-based methods: RealFusion \cite{realfusion}, Make-It-3D \cite{makeit3d}, and Magic123 \cite{magic123}. We replicated the implementations of these baseline methods by leveraging their publicly available open-source code and adhered strictly to their default parameter settings, eliminating any potential bias.

\begin{figure}
    \centering
    \includegraphics[width=\textwidth]{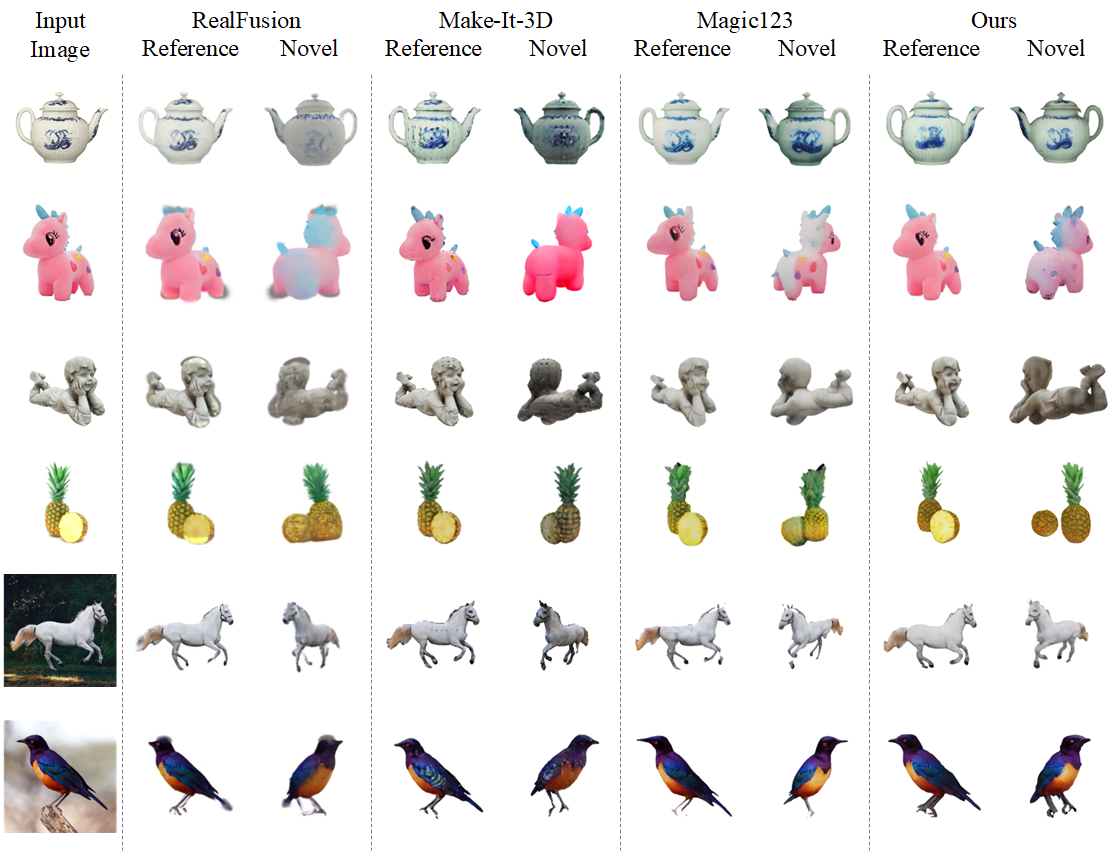}
    \caption{Qualitative comparison with RealFusion, Make-It-3D and Magic123 on synthetic and real-world images. Note that for real-world images, we first remove the background with \cite{u2net}, then use the preprocessed RGB-A images for 3D modeling. Due to the utilization of Multi-Word Textual Inversion, our method demonstrates better understanding of global semantic information in the image, \textit{e.g.}, color consistency, as shown in the second row, and spatial relationships, as shown in the fourth row.}
    \label{fig:qualitative}
\end{figure}

As shown in Fig. \ref{fig:qualitative}, each example is presented as a textured mesh rendered from the reference view and a novel view.
RealFusion demonstrates limitations in capturing the details of surface and overall texture quality, the reconstructed meshes also lack fidelity with the input.
While Make-It-3D exhibits some improvement in capturing finer surface details, it falls short of fully exploiting the information inherent in the input image due to the usage of a pretrained image caption model. 
Magic123, by leveraging supervision from both 2D and 3D diffusion priors, holds promise in generating novel views that are both visually and semantically consistent. 
However, it suffers from drawbacks associated with complicated training schemes and the imbalance between imposed constraints.
Both RealFusion and Magic123 rely on Marching Cubes to extract 3D meshes from trained NeRFs.
This introduces unwanted surface artifacts, resulting in unevenness within the reconstructed meshes.
On the contrary, MTFusion excels in capturing intricate geometric features and rich semantic information from single-view images (the second and fourth row in Fig. \ref{fig:qualitative}).
Furthermore, our method generates textures that are not only more photo-realistic but also exhibiting a higher degree of similarity to the original image, which ensures a visually appealing and informative representation of the reconstructed object.

\subsection{Quantitative Experiment}
Following \cite{magic123}, we use NeRF4 dataset and RealFusion15 dataset for quantitative analysis.
The NeRF4 consists 4 scenes collected from the synthetic NeRF dataset \cite{nerf}, while the RealFusion15 is composed of 15 real and synthetic images released by \cite{realfusion}.
We choose three metrics for evaluation, \textit{i.e.}, PSNR, LPIPS \cite{lpips}, and CLIP-similarity \cite{clip}. PSNR and LPIPS focus on measuring low-level and perceptual image similarity respectively, while CLIP-similarity assesses the semantic alignment between images and textual descriptions. 
As shown in Table \ref{tab:quantitative}, these metrics exhibit significant improvements over the baseline methods, underscoring the exceptional fidelity achieved by MTFusion.
Furthermore, the elevated CLIP-similarity score suggests a remarkable degree of 3D coherence between the reconstructed object and the reference view.
In simpler terms, the reconstructed object of our approach exhibits a high level of reality and consistency with the object depicted in the reference view, especially from a 3D perspective.
Additionally, due to the utilization of gradient-free optimization strategy, our approach achieves a 50\% speedup on 3D modeling, reducing the time for reconstructing a textured mesh from single image from about 90 minutes to 55 minutes.

\begin{table}
    \centering
    \caption{Quantitative results. The first three rows show the quantitative analysis of recent state-of-the-art methods. Rows 4-6 show the ablation results on multi-word textual inversion, with items in brackets indicating what type of word embedding is used. The last two rows show the ablation results on the enhanced FlexiCubes representation. Our framework employs the enhanced FlexiCubes and three types of word embedding, \textit{i.e.}, style, object and residual information.}
    \resizebox{0.9\linewidth}{!}{
    \begin{tabular}{lcccc}
        \toprule
         & PSNR$\uparrow$ & LPIPS$\downarrow$ & CLIP-similarity$\uparrow$ \\
        \midrule
        RealFusion & 14.36 & 0.23 & 0.58\\
        Make-It-3D & 13.98 & 0.19 & 0.63\\
        Magic123 & 19.87 & 0.17 & 0.77\\
        \midrule
        MTFusion (style) & 6.83 & 0.41 & 0.39\\
        MTFusion (object) & 17.88 & 0.24 & 0.71\\
        MTFusion (style \& object) & 20.14 & 0.16 & 0.73\\
        \midrule
        MTFusion (FlexiCubes) & 15.24 & 0.33 & 0.62\\
        MTFusion (ours) & 21.55 & 0.12 & 0.84\\
        \bottomrule
    \end{tabular}
    }
    \label{tab:quantitative}
\end{table}

\begin{figure}
    \centering
    \includegraphics[width=\textwidth]{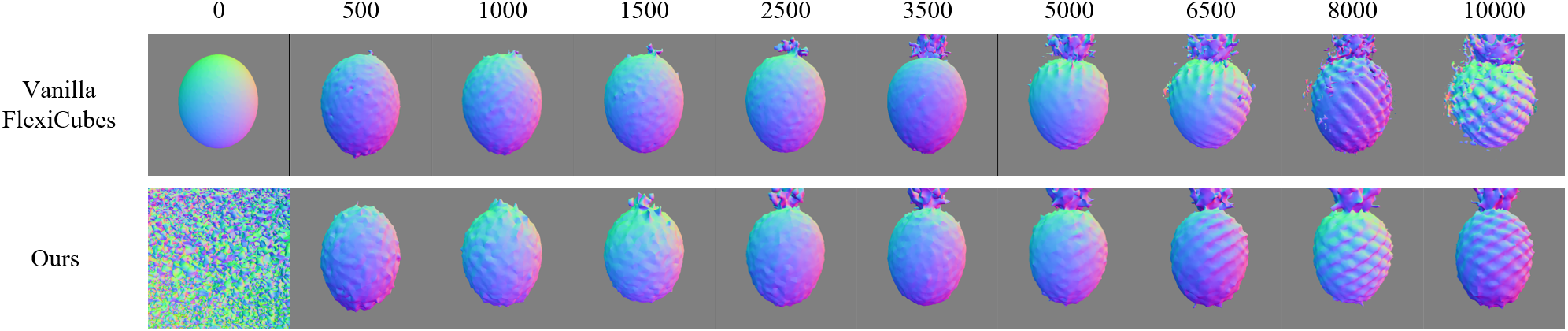}
    \caption{Ablation study on the enhanced FlexiCubes with hashgrid positional encoding. During the 3D mesh generation process from a given textual description ("A pineapple."), our enhanced FlexiCubes shows better training stability and robustness.}
    \label{fig:ablation}
\end{figure}

\subsection{Ablation Study}
\subsubsection{The Effect of Multi-Word Textual Inversion.}
To substantiate our findings, we conduct extensive experiments comparing the performance of single-word embeddings against their multi-word counterparts. 
While vanilla textual inversion remains a valuable tool for visual attribute extraction, the judicious adoption of multi-word embeddings promises to elevate the fidelity and authenticity of image-to-3D reconstruction.
As shown in Table \ref{tab:quantitative}, the results demonstrate that multi-word embeddings yield more accurate and nuanced representations, enhancing the fidelity of reconstructed 3D models.

\subsubsection{The Effect of Hashgrid Positional Encoding on FlexiCubes.}
To demonstrate the effectiveness of our approach in integrating FlexiCubes into the SDS-based training scheme, we conduct an ablation experiment on 3D shape generation from text only.
Vanilla FlexiCubes implements its 3D representation as parameters for direct optimization while we implicitly encode it into the SDF decoder network's parameters.
Note that vanilla FlexiCubes can be explicitly set as an ellipsoid, while our approach needs some warm-up (500 iterations) in the early training phase to get properly initialized.
Our analysis is based on the generation conditioned on a simple prompt, \textit{e.g.}, "A pineapple." 
We set both the coarse and fine training stages to 5000 iterations.
As shown in Table \ref{tab:quantitative} and Fig. \ref{fig:ablation}, our approach achieves more robust surface details and faster training speed than the standard version.
\section{Conclusion}
We propose MTFusion for tackling the challenge of reconstructing 3D models from single images. 
Unlike existing methods that capture limited semantic information, MTFusion leverages a novel multi-word textual inversion technique, which extracts a richer textual description encompassing the image's multi-perspective details.
For 3D representation, MTFusion utilizes FlexiCubes and enhances it using an SDF decoder network.
This network, designed for extracting the FlexiCubes parameters, accelerates training and improves the representation of intricate surface details in the final 3D model.
Evaluations demonstrate MTFusion's superiority over current methods in reconstructing high-quality 3D models from diverse synthetic and real-world images.
The effectiveness of MTFusion's design choices is further validated through ablation studies.

\section*{Acknowledgements}
This work is supported by the National Natural Science Foundation of China (No. 62176170, 61773270), and the Key Science and Technology Plans of Lhasa (No. LSKJ202306).

\bibliographystyle{splncs04}
\bibliography{ref}

\end{document}